\documentclass[letterpaper, 10 pt, conference]{ieeeconf}  
\IEEEoverridecommandlockouts                              
\usepackage{packages}

\newcommand{\vect}[1]{\boldsymbol{#1}}
\newcommand{\wrap}{\operatorname{wrap}}

\usepackage[utf8]{inputenc}
\usepackage{xcolor} 
\usepackage[most]{tcolorbox} 

\usepackage{packages}

\definecolor{myPurple}{RGB}{191, 179, 219}

\newtcolorbox{infoBubble}[1]{
  colback         = myPurple!15!white, 
  colframe        = myPurple!80!black, 
  fonttitle       = \bfseries,         
  colbacktitle    = myPurple!80!black, 
  coltitle        = white,             
  rounded corners,                     
  arc             = 3mm,               
  boxrule         = 1pt,               
  title           = {#1},              
  halign title    = left,              
  attach boxed title to top left={yshift=-0.5mm, xshift=2mm},
  boxed title style={
    arc=2mm, 
  }
}

\title{\LARGE\bf

RAVEN: Reinforcement-Adaptive Visibility-Graph Planning for Robust Humanoid Navigation with Collision-Free MPC

}

\author{Ruochen Hou$^{1}$, Shiqi Wang$^{1}$, Beom Jun Kim$^{1}$, Hanzhang Fang$^{1}$, Mehak Singal$^{1}$, and Dennis W.~Hong$^{1}$
\thanks{
$^{1}$Robotics and Mechanisms Laboratory (RoMeLa), Department of Mechanical and Aerospace Engineering, University of California, Los Angeles, CA 90095, USA.
        {\tt\small \{houruochen, edmond.wang, bj007, hfang09, mehaksingal, dennishong\}@ucla.edu}}
}

\begin{document}
\maketitle
\thispagestyle{empty}
\pagestyle{empty}

\begin{abstract}
Humanoid navigation in dynamic environments requires long-horizon planning while respecting short-horizon dynamic and safety constraints. Classical visibility-graph planners combined with model predictive control (MPC) can efficiently generate collision-free trajectories, but their performance depends on manually tuned parameters and accurate system modeling. In real robotic systems, control delays, state-estimation noise, and locomotion uncertainties can cause overshoot and constraint violations even when the nominal path is geometrically optimal. We propose \textbf{RAVEN}, a hierarchical reinforcement learning (RL)–MPC framework for robust humanoid navigation. Unlike prior approaches that use learning to tune cost weights or replace planning entirely, RAVEN employs RL to adapt the \emph{geometric construction} of a visibility-graph planner by modifying obstacle inflation and related graph parameters. By directly reshaping the free-space geometry, the learned planner alters the topology of the global path to compensate for delay and tracking imperfections. A collision-free MPC layer then tracks the planned trajectory while explicitly enforcing velocity bounds and obstacle-avoidance constraints. By training under realistic delays and observation noise, RAVEN learns planning adaptations that improve robustness while retaining explicit long-horizon geometric planning and constrained optimization, in contrast to end-to-end learning approaches. We evaluate RAVEN against a manually tuned visibility-graph MPC baseline and a pure RL navigation policy. Results demonstrate reduced overshoot near obstacles, improved robustness in narrow passages, and more reliable navigation under delay and noise. These findings indicate that reinforcement-adaptive graph construction combined with constrained MPC provides an effective and interpretable alternative to end-to-end learning for robust humanoid navigation.
\end{abstract}

\section{Introduction}

The promise of humanoid robotics lies in their potential to navigate the complex, vertical, and unstructured environments designed for humans. However, realizing this agility in the real world exposes a fundamental friction between idealized geometric planning and unordered dynamic execution. Classical navigation stacks often pair a global planner, such as a visibility graph, with a local Model Predictive Control (MPC) tracking layer. While geometrically optimal in theory, these pipelines assume near-perfect system modeling. In reality, deploying these systems on physical humanoids introduces control delays, state-estimation noise, and locomotion uncertainties. When a robot inevitably lags behind its nominal path, a trajectory that perfectly grazes an obstacle in simulation can quickly devolve into suboptimal repetitions near the obstacle boundary.


To address navigation under these real-world uncertainties, the field has largely bifurcated into two distinct paradigms. 

\begin{figure} [h]
    \centering
    \includegraphics[width=0.98\linewidth]{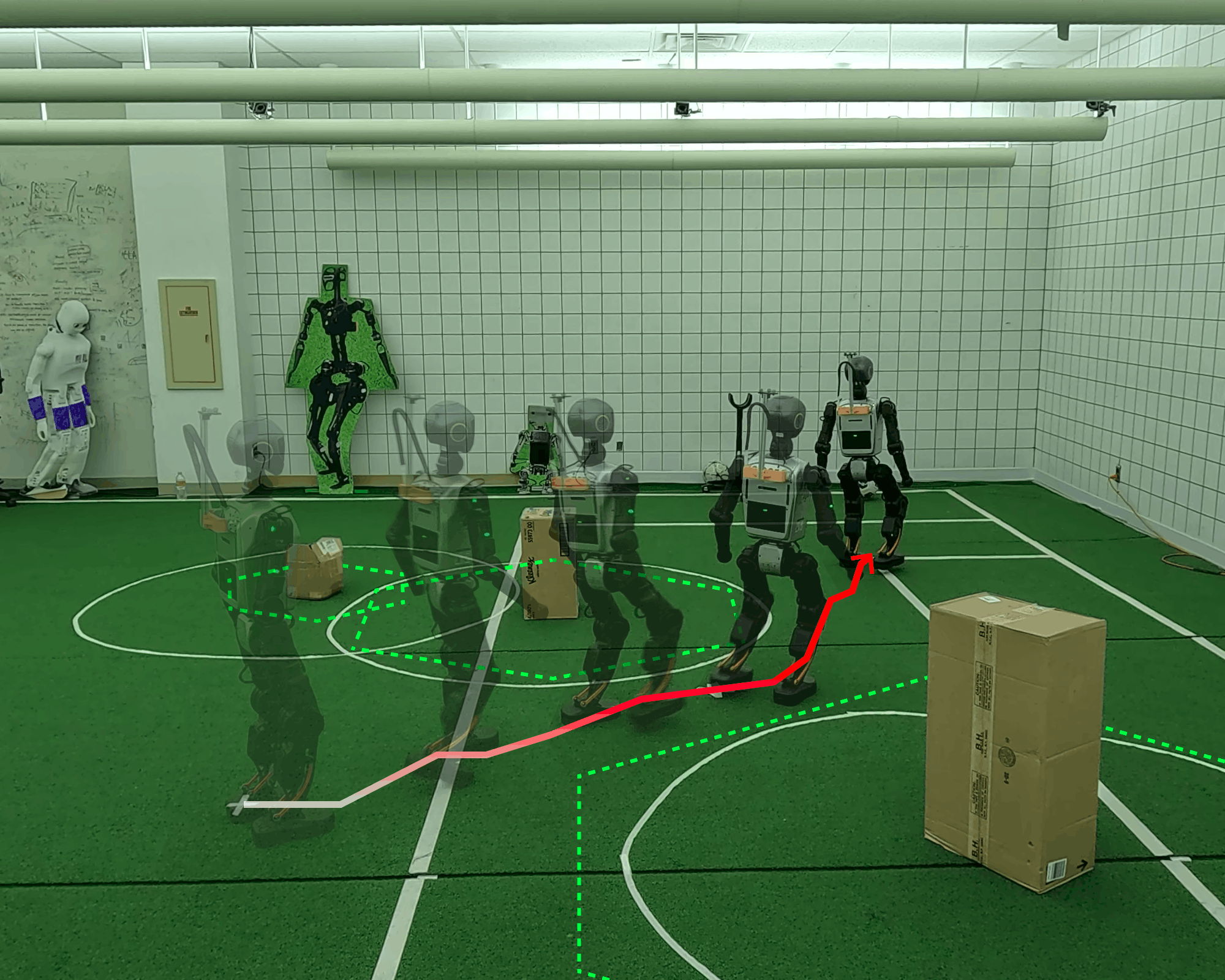}
    \caption{The RAVEN framework deployed on the T1 Booster biped navigating among obstacles in a half-sized RoboCup soccer field. Hexagons indicate the adaptive obstacle inflation regions, whose radii are determined by the RL meta policy to account for control delays and tracking uncertainties.}
    \label{fig:field_test}
\end{figure}

On one hand, end-to-end Deep Reinforcement Learning (DRL) approaches, such as \cite{compass_2025, anymal_parkour_2023, robot_parkour_learning_2023}, train monolithic neural networks to map raw perception directly to motor commands. While these methods exhibit remarkable agility and emergent skills, they suffer from a severe lack of interpretability, making it nearly impossible to formally verify safety constraints or diagnose failures in critical scenarios. Furthermore, while heavily utilized on quadrupedal platforms - which are inherently statically stable and less prone to hardware damage - humanoid robots demand stricter, mathematically guaranteed safety margins alongside agile navigation. 


On the other hand, classical graph-search and MPC-based planning frameworks \cite{davg-cfmpc, shirai2021lto} provide interpretable long-horizon planning, explicit constraint handling, and predictable behavior. However, the real-world efficacy is bottlenecked by static, manually tuned geometric heuristics - most notably, obstacle inflation radii. A fixed inflation parameter cannot universally account for varying speeds, narrow passages, or fluctuating system delays, forcing a painful compromise between conservative sluggish navigation and aggressive, collision-prone trajectories. Neither extreme is sustainable for robust deployment.


Bridging this divide requires a hybrid architecture that leverages the complementary strengths of both methods without inheriting their brittleness. While recent works have explored hierarchical integration - such as using RL to predict discrete subgoals or employing heuristics to toggle between classical planners - these methods fundamentally treat the underlying global planner's geometric rules as a rigid, unalterable module. Other approaches seek deeper integration by using RL to tune the algebraic cost weights of the local MPC, sometimes embedding the solver entirely within a differentiable computational graph. However, differentiable optimization introduces severe computational overhead, and adjusting penalty matrices provides an indirect, mathematically opaque mechanism for reshaping a robot's spatial behavior. A more direct, interpretable, and computationally efficient direction is to use learning not to tune the local algebraic controller, but to actively adapt the geometric optimization landscape of the global planner itself.



\begin{figure} [!t]
    \centering
    \includegraphics[width=0.98\linewidth]{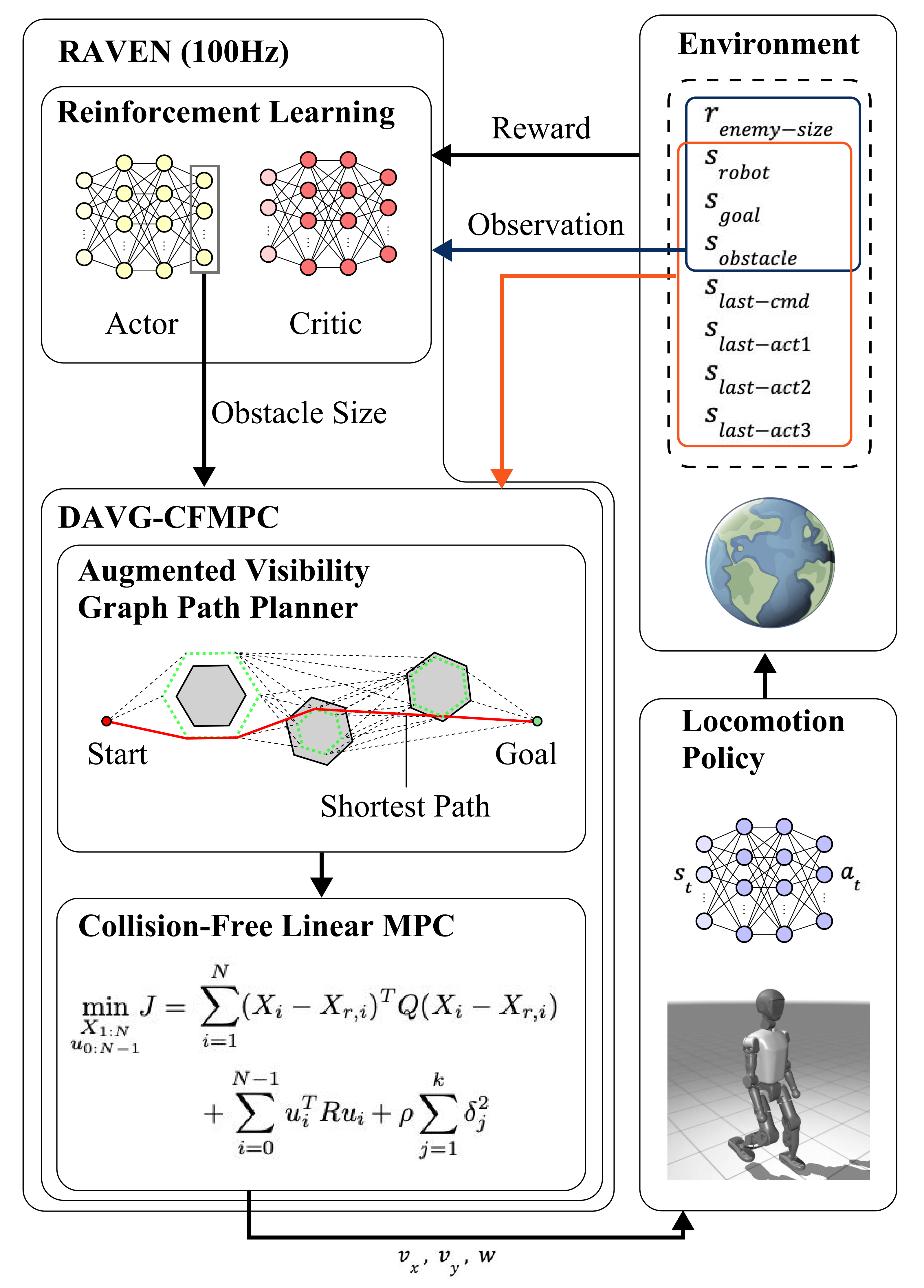}
    \caption{Overview of the RAVEN architecture. The framework consists of three hierarchical layers: (1) an actor-critic RL meta policy that adapts the obstacle inflation parameters of the visibility graph, (2) a DAVG-cfMPC planner that computes the shortest collision-free path and tracks it via linear MPC, and (3) a locomotion policy that converts velocity commands into joint-level control. The RL meta policy receives observations and rewards from the environment to continuously adapt planning parameters during execution.}
    \label{fig:overall_archi}
\end{figure}

In this work, we propose \textbf{RAVEN} (Reinforcement-Adaptive Visibility-Graph Planning with Collision-Free MPC), a hierarchical framework for agile humanoid navigation. In RAVEN, a reinforcement learning agent operates at a high level to adapt the construction of a visibility-graph planner rather than directly generating control commands. This allows the robot to effectively ``learn how to plan,'' continuously adjusting the geometric structure of the planning graph to balance velocity, safety margins, and collision avoidance according to system characteristics such as locomotion dynamics, control delays, and perception noise. The resulting trajectory is then executed by a collision-free MPC layer that enforces dynamic feasibility and safety constraints during motion execution. Furthermore, because low-level motion execution and constraint enforcement are handled by MPC rather than a learned policy, the proposed framework relies less on precise simulation of robot dynamics, which can reduce the sim-to-real gap compared with end-to-end RL navigation approaches.

The main contributions of this paper are as follows:

\begin{enumerate}
\item \textbf{Reinforcement-adaptive visibility-graph planning.}
We propose a learning-augmented visibility-graph planner in which reinforcement learning adapts the geometric construction of the graph by modifying obstacle inflation and related planning parameters. Unlike prior approaches that tune cost weights, this directly reshapes the free-space geometry and alters the topology of the resulting global path.

\item \textbf{A hierarchical RL--MPC navigation framework.}
We introduce RAVEN, an architecture that combines RL-adaptive visibility-graph planning for global path generation with collision-free model predictive control (cf-MPC) for local trajectory tracking. The MPC layer explicitly enforces velocity limits, acceleration limits, and obstacle-avoidance constraints while following the RL-adapted plan, preserving explicit geometric planning and transparent constraint enforcement.

\item \textbf{Guided exploration through shortest-path structure.}
Because the visibility graph computes shortest paths under the current obstacle geometry, the RL adaptation effectively searches over a family of shortest-path solutions. This structured prior guides exploration toward near-optimal navigation strategies and reduces convergence to suboptimal behaviors compared with pure reinforcement learning.

\item \textbf{Robust navigation under delay and observation noise.}
By training under realistic control delays and perception noise, the learned planner adapts obstacle inflation and path geometry to mitigate overshoot and tracking errors. Experiments in simulation and hardware demonstrate improved reliability in challenging scenarios such as narrow passages and delays, while showing consistent performance across simulation and real-world deployments.


\end{enumerate}
\section{Related Work} 

\subsection{End-to-End Learning for Agile Locomotion}
\label{sec:End-to-End_method}
The dominant trend in modern legged robotics has been the shift toward end-to-end Deep Reinforcement Learning (DRL) policies that map raw sensor observations directly to joint commands. These monolithic architectures have demonstrated remarkable capabilities in traversing unstructured terrain. For instance, \cite{robot_parkour_learning_2023} proposes a vision-based policy capable of executing diverse skills - such as climbing and leaping - emerging entirely from a unified neural network without explicit reference motions. Similarly, the \cite{compass_2025} framework introduces a cross-embodiment pipeline that distills specialist policies into a single generalist agent, using residual RL to adapt a foundational navigation policy to different robot morphologies. Other frameworks adopt a hierarchical but fully learned architecture; \cite{anymal_parkour_2023} decomposes the problem into a low-level library of specialized locomotion skills and a high-level navigation policy that selects the appropriate skill based on perception.

While these learning-based methods excel at agility and generalization, they often function as "black boxes". For bipedal humanoids, which possess complex, underactuated dynamics, the lack of interpretability presents a challenge for formal verification. Ensuring strict dynamic constraint satisfaction and guaranteed safety margins remain difficult with purely neural architectures, motivating the need for hybrid methods that combine learned agility with explicit mathematical safety bounds. 



\subsection{RL-Driven Path Generation and Subgoal Planning}


While pure reinforcement learning policies can be formulated to map observations directly to navigation commands, addressing long-horizon planning in complex environments often motivates a shift towards hierarchical frameworks. In these architectures, RL typically acts as a high-level waypoint or subgoal generator for a classical low-level controller.

Recent works have demonstrated the efficacy of this partitioned architecture. For example, \cite{dynamic_subgoal_2025} and \cite{hierarchical_crowd_2025} employ RL and Graph Neural Networks, respectively, to dynamically predict discrete local subgoals for an MPC to track through cluttered or dynamic environments. Similarly, \cite{rl_path_gen_2023} utilizes a deep Markov model to generate navigation waypoints in unknown environments without prior maps. While these approaches successfully blend learned spatial intuition with the stable locomotion of model-based control, they fundamentally treat the underlying path generation as a point-to-point prediction task. The global planner's geometric rules remain rigid and inaccessible to the learning agent. RAVEN departs from this paradigm by using RL not to predict arbitrary waypoints, but to systematically alter the geometric construction of the visibility graph itself. By dynamically adapting obstacle inflation, RAVEN explicitly reshapes the global topology to account for the execution delays and tracking errors of the downstream MPC. 

\subsection{Synergies of RL and MPC}

To combine the adaptability of learning with the mathematical rigor of control theory, recent research has explored various integrations of RL and Model Predictive Control (MPC). These hybrid architectures can be broadly categorized by the structural relationship between the learning and control modules.


\subsubsection{Action-level Integration and Imitation}

Rather than integrating the planning processes, several frameworks combine RL and classical control at the policy or execution level. Residual learning frameworks, such as \cite{residual_rl}, utilize RL to output an additive continuous correction to a nominal hand-crafted controller, retaining the base controller's stability while extending its dynamic range. Conversely, works such as \cite{MPC-Net} and \cite{MPC-guided_policy_search} invert this relationship via imitation learning. In these approaches, a computationally expensive, state-based MPC acts as an expert demonstrator to train a faster, reactive neural network policy, effectively distilling the MPC's optimization into a sensorimotor mapping. These methods tend to treat the controller as a fixed module to be patched or imitated. 




\subsubsection{Gating Mechanisms and Safety Filters}

Other architectures maintain distinct classical and learning-based modules, managing their interaction through high-level gating or filtering mechanisms. Planner switching methods \cite{hybrid_planner_2025} employ heuristics to toggle execution between a classical planner (e.g., for stable, static environments) and an RL policy (e.g., for reactive, dynamic obstacle avoidance). In safety-critical contexts, frameworks like Model Predictive Shielding \cite{model_predictive_shielding} employ the MPC strictly as a safeguard. In these setups, an explicit model-based safety filter evaluates the action proposed by a primary RL agent, intervening or overriding the command only if the action is predicted to violate strict kinematic or collision constraints. While effective for preventing catastrophic failure, these reactive architectures can result in overly conservative or disjointed behavior. 



\subsubsection{Learning System Dynamics for MPC}
A parallel line of work enhances the MPC's internal optimization by using neural networks to approximate complex or highly nonlinear system dynamics. Frameworks such as \cite{real-time_neural_MPC} integrate a learned neural dynamics model directly into the MPC optimization loop, computing local linear approximations of the network to serve as dynamic constraints. Other methods use learning to quantify execution uncertainty; for instance, Dynamic Tube MPC \cite{dynamic_tube_mpc} leverages massively parallel simulation to learn a dynamic error representation. This model characterizes the expected low-level tracking error as a function of the planned trajectory, allowing the MPC to optimize a path while strictly ensuring the learned "tube" of potential tracking deviation remains in collision-free space. While these approaches improve localized tracking precision, they still rely on a nominal geometric path that may be topologically suboptimal under system delays. RAVEN shifts this burden to the high-level planner, ensuring the requested path inherently accounts for these uncertainties before the local MPC begins its optimization. Furthermore, introducing neural-network-based system dynamics, which results in a nonlinear model, significantly increases the computational cost of the optimization. Consequently, Dynamic Tube MPC can only run at around 10 Hz.

\subsubsection{RL-tuned and differentiable MPC}

Closest to the proposed RAVEN framework are methodologies that utilize RL to actively tune the internal parameters of the MPC. Early instantiations demonstrated the ability to automate the search for optimal, static cost weights via iterative offline simulations \cite{automated_tuning}. To handle varying environmental conditions, adaptive approaches \cite{weights_varying_mpc} were developed to dynamically select from a discrete set of pre-optimized weight vectors based on current state observations.

The most advanced integration involves end-to-end differentiability. Notably, Actor-Critic MPC \cite{actor_critic_MPC} embeds a differentiable MPC solver as the final layer of the actor network. This architecture allows the neural policy to map state observations directly to the parameters of the MPC's quadratic cost matrices, optimizing the local behavior online. While highly expressive, differentiable MPC frameworks are notoriously computationally intensive, requiring the backpropagation of gradients through a complex trajectory solver at every step. Furthermore, adjusting algebraic cost weights is an indirect mechanism for altering a robot's spatial behavior. RAVEN bypasses these computational bottlenecks by tuning the \textbf{geometric parameters} of a high-level visibility graph - specifically, obstacle inflation - rather than the algebraic weights of the local controller. This explicitly reshapes the topological free-space to account for execution delays, allowing a standard, non-differentiable MPC to track a dynamically robust global path.


\section{The RAVEN Framework} 

In this section, we present the proposed \textbf{RAVEN} framework, a hierarchical architecture designed for robust humanoid navigation. As shown in \cref{fig:overall_archi}, the system is organized into three layers. At the highest level, a reinforcement learning (RL) meta-policy adapts the construction parameters of the visibility-graph planner, enabling the robot to adjust planning strategies according to environmental conditions and system dynamics. At the intermediate level, a visibility-graph planner generates a global path, while a collision-free model predictive controller (cf-MPC) tracks the planned trajectory under explicit dynamic and safety constraints. Finally, at the lowest level, a locomotion policy executes the commanded velocity to produce stable humanoid motion. This hierarchical structure separates long-horizon planning, trajectory optimization, and low-level control, allowing learning to adapt the planning process while preserving the safety and interpretability of classical planning and control methods.

\subsection{DAVG-cfMPC Formulation}

We build upon the Dynamic Augmented Visibility Graph (DAVG) and Collision-Free MPC (cf-MPC) pipeline \cite{davg-cfmpc}, a system previously validated for real-time humanoid navigation. Since the full algorithmic formulation is detailed in prior work, we focus specifically on the geometric parameters modulated by our learning framework.

DAVG computes a kinematically feasible global path by discretizing the environment into a visibility graph, routing the robot around obstacles that are artificially enlarged by a safety margin. This trajectory serves as the reference for the local cf-MPC, which tracks the path while enforcing dynamic feasibility. Crucially, cf-MPC formulates collision avoidance as a linearized soft constraint within a convex Quadratic Program (QP). For each obstacle $j$, the constraint is linearized around the reference trajectory:

$$[x_k-x_{obs,j}, y_k-y_{obs, j}]\cdot V_{k,j} \ge \|V_{k,j}\|(R_{obs,j} - \delta_j)$$

where $V_{k,j}$ is the relative position vector between the robot and the obstacle, $\delta_j\ge 0$ is a slack variable to ensure QP feasibility, and $R_{obs,j}$ is the obstacle inflation radius.

\subsection{RL Meta Policy}

At the highest level of the proposed framework, a reinforcement learning (RL) meta policy is introduced to adapt the global planning behavior. Unlike end-to-end RL navigation policies that directly output motion commands, the meta policy in RAVEN does not control the robot velocity. Instead, it outputs planning parameters that modify the geometric construction of the visibility graph used by the path planner.

Specifically, the meta policy adjusts obstacle inflation parameters that determine the effective size of obstacles in the planning space. By modifying these parameters, the RL agent can reshape the free-space geometry and influence the topology of the resulting visibility graph. This allows the system to adapt the planned path to account for system characteristics such as locomotion dynamics, control delays, and perception noise. For example, increasing obstacle inflation can create safer paths with larger clearance, while smaller inflation may allow more aggressive trajectories through narrow passages.

The meta policy observes the robot state, goal information, obstacle locations, and selected planning-related variables from the previous timestep. Based on these observations, it outputs a set of parameters that determine the obstacle inflation values used for graph construction. The resulting visibility graph is then passed to the intermediate planning layer, where the shortest path is computed and tracked by the collision-free MPC controller.

By restricting the RL policy to modify planning parameters rather than directly controlling the robot, the proposed approach preserves the structure and interpretability of classical planning methods while allowing the system to adapt to uncertainties and model mismatches.





\subsubsection{Observation Space}

The observation space of the meta policy includes robot state information, goal-related features, obstacle locations, and several planning-related variables, as summarized in Table~\ref{tab:meta_policy_obs}. The robot pose in the world frame is represented by its planar position and orientation using the representation $[x,y,\sin\psi,\cos\psi]$, where the sine and cosine encoding avoids discontinuities associated with angular wrap-around. Similarly, the goal pose is provided both in the world frame and in the robot body frame, where the body-frame representation captures the relative goal direction with respect to the robot. 

To describe the surrounding environment, obstacle positions are included in both world and body frames. Each obstacle is represented only by its planar position $[o_x,o_y]$ without orientation, since obstacles are modeled as circular regions in the navigation space. Including both coordinate frames allows the policy to reason about the global geometry of the environment as well as the relative spatial relationships between the robot and nearby obstacles.

In addition to the geometric observations, the policy also receives planning-related variables from the previous timestep. These include the previous MPC velocity command $\vect{u}_{t-1}^{\text{cmd}}$ and the previous meta action $a_{t-1}$ that controls obstacle inflation in the visibility graph. Providing these variables helps the policy maintain temporal consistency when adapting the planning parameters.

The policy is trained using an asymmetric actor--critic architecture. The actor receives observations that may contain realistic perception noise and control delay, while the critic receives privileged observations that concatenate both the noisy/delayed state and the clean state. This design improves value estimation during training while ensuring that the deployed policy relies only on observations available in the real system.
The details of observation are summarized in \cref{tab:meta_policy_obs}.

\begin{table}[t]
\centering
\caption{Observation components used by the RAVEN meta RL policy. Actor observations correspond to noisy and delayed robot states, while the critic receives privileged observations including both noisy and clean states. 
Robot and goal poses are represented as $[x,y,\sin\psi,\cos\psi]$, and each obstacle is represented by its position $[o_x,o_y]$. 
In the default configuration $K=3$, giving $d=30$ and a critic input dimension of $60$.}
\label{tab:meta_policy_obs}

\footnotesize
\begin{tabular}{l c c c}
\toprule
\textbf{Observation component} & \textbf{Dim.} & \textbf{Actor} & \textbf{Critic} \\
\midrule
Robot pose (world) & 4 & \checkmark & \checkmark \\
Goal pose (world) & 4 & \checkmark & \checkmark \\
Goal pose (body frame) & 4 & \checkmark & \checkmark \\
Obstacle positions (world) & $2K$ & \checkmark & \checkmark \\
Obstacle positions (body frame) & $2K$ & \checkmark & \checkmark \\
Previous MPC command & 3 & \checkmark & \checkmark \\
Previous meta action & $K$ & \checkmark & \checkmark \\
\midrule
Total actor observation dimension $d$ & $15+5K$ &  &  \\
Privileged critic observation & $2d$ &  & \checkmark \\
\bottomrule
\end{tabular}

\vspace{0.3em}
\end{table}

\subsubsection{Action Space}

The meta policy outputs a continuous action vector
\[
a_t \in [-1,1]^K ,
\]
where each element corresponds to one obstacle in the environment. 
Each action component is interpreted as a parameter controlling the effective obstacle size used during visibility-graph construction. 
Specifically, the action is mapped to a planning radius for each obstacle through an affine transformation

\[
r_{t,i} = \frac{1}{2}(a_{t,i}+1)(r_{\max}-r_{\min}) + r_{\min}, 
\quad i=1,\dots,K ,
\]

where \(r_{\min}\) and \(r_{\max}\) define the allowable range of obstacle inflation. 
These radii are then used to parameterize the obstacle geometry in the visibility-graph planner, effectively modifying the free-space geometry and influencing the resulting path topology.

\begin{figure}[htbp]
    \centering
    
    \begin{subfigure}{0.23\textwidth}
        \centering
        \includegraphics[width=\linewidth]{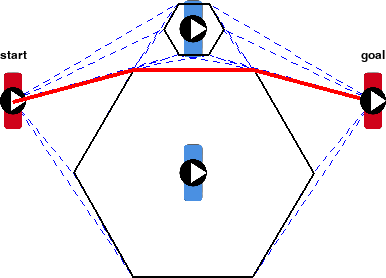}
        \caption{shift path up}
    \end{subfigure}
    \hfill
    \begin{subfigure}{0.23\textwidth}
        \centering
        \includegraphics[width=\linewidth]{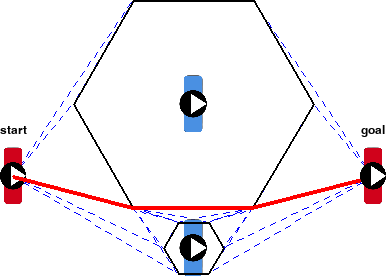}
        \caption{shift path down}
    \end{subfigure}

    \vspace{0.2cm}

    \begin{subfigure}{0.23\textwidth}
        \centering
        \includegraphics[width=\linewidth]{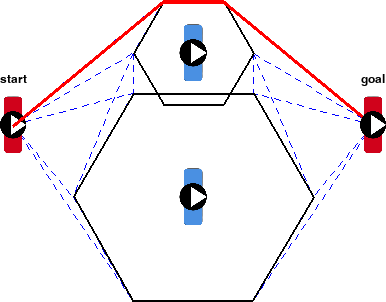}
        \caption{alternative path up}
    \end{subfigure}
    \hfill
    \begin{subfigure}{0.23\textwidth}
        \centering
        \includegraphics[width=\linewidth]{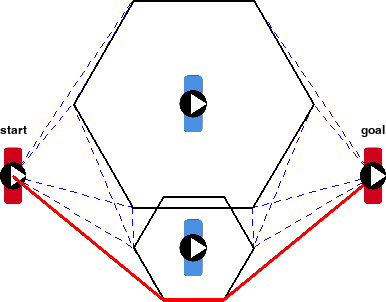}
        \caption{alternative path down}
    \end{subfigure}

    \caption{Illustration of how adapting obstacle size changes the visibility graph and alters the resulting path. The planned path is shown as a red line, and the edges of the visibility graph are shown as blue dashed lines.}
    \label{fig:adaptive_visibility_graph}
\end{figure}

As shown in \cref{fig:adaptive_visibility_graph}, by increasing the planning radius, the meta policy can enlarge obstacle regions and generate trajectories with larger safety margins, which can improve robustness to delays and tracking errors. Conversely, reducing the radius allows the planner to produce shorter paths that pass closer to obstacles, enabling more aggressive navigation when conditions permit. The resulting planning parameters are passed to the visibility-graph planner, whose output trajectory is subsequently tracked by the collision-free MPC layer.

\subsubsection{Reward Function}

The reward function is designed to encourage efficient and safe navigation while maintaining smooth planning behavior of the meta policy. The overall reward is composed of several components summarized in Table~\ref{tab:raven_reward}. 

A time penalty and a path-length penalty are introduced to encourage the robot to reach the goal efficiently and avoid unnecessarily long trajectories. These terms discourage the planner from generating overly conservative paths that significantly increase travel distance. 

To promote safety, collision and penetration penalties are applied when the robot approaches or intersects with obstacle regions. The collision term penalizes configurations where the robot enters a predefined safety radius around obstacles, while the penetration term provides an additional penalty proportional to the amount of overlap. These terms encourage the meta policy to select obstacle inflation parameters that generate collision-free paths with sufficient clearance.

An action-rate penalty is included to regularize the meta policy output. This term penalizes large changes in obstacle inflation parameters between consecutive timesteps, promoting smoother adaptation of the planning parameters and preventing unstable planning behavior.

Finally, sparse terminal rewards are used to represent task success and failure. A large positive reward is granted when the robot reaches the goal pose within specified position and orientation tolerances, while a large negative penalty is applied if the robot falls during execution. These terminal rewards ensure that successful navigation is strongly favored during training. 
The reward details are summarized in \cref{tab:raven_reward}.

\begin{table}[t]
\centering
\caption{Summary of reward function used by the RAVEN meta RL policy.}
\label{tab:raven_reward}

\scriptsize
\begin{tabular}{>{\raggedright\arraybackslash}p{0.24\linewidth} >{\raggedright\arraybackslash}p{0.40\linewidth} c}
\toprule
\textbf{Component} & \textbf{Equation} & \textbf{Weight} \\
\midrule

Time term &
$1$ &
$w_{\text{time}}=-5.0$ \\

Path length &
$\|\vect{p}_t-\vect{p}_{t-1}\|_2$ &
$w_{\text{path}}=-20.0$ \\

Collision penalty &
$\sum_i \mathbf{1}\{\|\vect{p}_t-\vect{o}_i\|_2 < r_{\text{pen}}\}$ &
$w_{\text{col}}=-8.0$ \\

Penetration penalty &
$\sum_i \max(0, r_{\text{pen}}-\|\vect{p}_t-\vect{o}_i\|_2)$ &
$w_{\text{inside}}=-80.0$ \\

Action rate &
$\sum_i (r_{t,i}-r_{t-1,i})^2$ &
$w_{\text{ar}}=-0.5$ \\

Success bonus &
$\mathbf{1}\{\|\vect{p}_t-\vect{g}\|_2<\epsilon_p
\land |\wrap(\psi_t-\psi_g)|<\epsilon_\psi\}$ &
$R_{\text{succ}}=5000$ \\

Fall penalty &
$\mathbf{1}\{\text{fall detected}\}$ &
$R_{\text{fall}}=-50000$ \\

\bottomrule
\end{tabular}
\end{table}

\subsubsection{Episode Design}

Each training episode consists of a navigation task where the robot starts from a randomly initialized pose and must reach a specified goal pose while avoiding obstacles. The episode terminates when the goal is reached, a failure condition occurs, or a maximum time horizon is exceeded.

\section{Experimental Setup}

\subsection{Baseline Methods}

We consider two baseline methods in this paper: an MPC baseline and an end-to-end RL baseline. The MPC baseline uses the DAVG--cfMPC method \cite{davg-cfmpc} with a fixed obstacle radius of 1~m. 

For the end-to-end RL baseline, the observation space is summarized in \cref{tab:obs_direct}, and the reward function is summarized in \cref{tab:reward_direct}. The reward design combines several components described in \cref{sec:End-to-End_method} to balance path efficiency and collision avoidance.

All other settings, including velocity limits, acceleration limits, the robot model, and the injected delay and observation noise, are kept identical across the three methods to ensure a fair comparison.

\begin{table}[t]
\centering
\caption{Observation components used by the end-to-end RL baseline. Actor observations correspond to noisy and delayed robot states, while the critic receives privileged observations including both noisy and clean states. Robot and goal poses use the representation $[x,y,\sin\psi,\cos\psi]$, while each obstacle is represented by its position $[o_x,o_y]$. In the default configuration $K=3$, giving $d=27$ and a critic input dimension of $54$.}
\label{tab:obs_direct}

\footnotesize
\begin{tabular}{l c c c}
\toprule
\textbf{Observation component} & \textbf{Dim.} & \textbf{Actor} & \textbf{Critic} \\
\midrule
Robot pose (world) & 4 & \checkmark & \checkmark \\
Goal pose (world) & 4 & \checkmark & \checkmark \\
Goal pose (body frame) & 4 & \checkmark & \checkmark \\
Obstacle positions (world) & $2K$ & \checkmark & \checkmark \\
Obstacle positions (body frame) & $2K$ & \checkmark & \checkmark \\
Previous velocity command & 3 & \checkmark & \checkmark \\
\midrule
Total actor observation dimension $d$ & $15+4K$ &  &  \\
Privileged critic observation & $2d$ &  & \checkmark \\
\bottomrule
\end{tabular}
\end{table}

\begin{table}[t]
\centering
\caption{Summary of reward function used by the end-to-end RL baseline. Default parameters are $\epsilon_p=0.2$, $\epsilon_\psi=0.2$, and $r_{\text{pen}}=1.0$.}
\label{tab:reward_direct}

\scriptsize
\begin{tabular}{>{\raggedright\arraybackslash}p{0.24\linewidth} >{\raggedright\arraybackslash}p{0.40\linewidth} c}
\toprule
\textbf{Component} & \textbf{Equation} & \textbf{Weight} \\
\midrule

Position error &
$\|\vect{p}_t-\vect{g}\|_2$ &
$s_{\text{pos}}=-2.0$ \\

Yaw error &
$|\wrap(\psi_t-\psi_g)|$ &
$s_{\text{yaw}}=-0.5$ \\


Collision penalty &
$\sum_i \mathbf{1}\{\|\vect{p}_t-\vect{o}_i\|_2 < r_{\text{pen}}\}$ &
$s_{\text{col}}=-5.0$ \\

Penetration penalty &
$\sum_i \max(0,\, r_{\text{pen}}-\|\vect{p}_t-\vect{o}_i\|_2)$ &
$s_{\text{inside}}=-5.0$ \\

Progress shaping &
$d_{t-1}-d_t,\;\; d_t=\|\vect{p}_t-\vect{g}\|_2$ &
$s_{\text{prog}}=0.5$ \\

Smoothness penalty &
$\|\vect{u}_t^{\text{cmd}}-\vect{u}_{t-1}^{\text{cmd}}\|_2^2$ &
$s_{\text{smooth}}=-0.1$ \\


Forward motion &
$\begin{aligned}[t]
(\vect{p}_t-\vect{p}_{t-1})^\top
\frac{\vect{g}-\vect{p}_t}{\|\vect{g}-\vect{p}_t\|}
\end{aligned}$ &
$s_{\text{fwd}}=0.5$ \\

Time term &
$1$ &
$s_{\text{time}}=-5.0$ \\

Success bonus &
$\mathbf{1}\{\|\vect{p}_t-\vect{g}\|_2<\epsilon_p
\land |\wrap(\psi_t-\psi_g)|<\epsilon_\psi\}$ &
$s_{\text{succ}}=100.0$ \\


Fall penalty &
$\mathbf{1}\{\text{inner env done}\}$ &
$s_{\text{fall}}=-50000$ \\

\bottomrule
\end{tabular}
\end{table}

\subsection{RL Training Details}

Both the proposed RAVEN framework and the pure RL baseline are trained using
Proximal Policy Optimization (PPO) implemented in the Brax training framework.
The policy and value networks are multilayer perceptrons with hidden sizes
$(512, 256, 128)$ and Swish activation. The policy outputs a tanh-squashed
diagonal Gaussian distribution. Actor observations correspond to the noisy
robot state, while the critic receives privileged observations that include
both noisy and clean states to stabilize value estimation.

Training is performed in JAX using the MuJoCo Playground MJX environments,
which enable fully vectorized physics simulation on GPU. For the RAVEN method,
the DAVG--cfMPC layer in RAVEN is implemented using JAX and solved via projected
gradient optimization using JAXopt. This allows both the physics simulation
and the MPC optimization to run on the GPU and be executed in parallel across
multiple environments.

For RL+MPC training, we use $N_{\text{env}}=1024$ parallel environments with an
unroll length of $T=32$. Due to the embedded MPC optimization, the training
throughput is approximately $10\,000$ simulation steps per second (SPS) on an
NVIDIA RTX~5090 GPU. In comparison, the pure RL baseline achieves roughly
$100\,000$ SPS because it does not require solving an optimization problem at
each control step. The computational bottleneck of RAVEN therefore arises from
the MPC optimization rather than the physics simulation itself.

To ensure a fair comparison between methods, we allocate comparable total
training time for both approaches. Specifically, the RAVEN policy is trained
for $10^8$ environment steps, while the pure RL baseline is trained for
$10^9$ steps due to its higher simulation throughput. This protocol ensures
that the RL baseline is not disadvantaged by shorter training duration despite
its faster simulation speed.

Additional PPO settings follow standard Brax configurations, including
generalized advantage estimation ($\lambda=0.95$), clipping parameter
$\epsilon=0.2$, advantage normalization, observation normalization, and
gradient norm clipping.


\section{Results}
\label{sec:results}
\subsection{Simulation Results: Robustness to Delay and Noise}

\begin{figure}[htbp]
    \centering
    
    \begin{subfigure}{0.23\textwidth}
        \centering
        \includegraphics[width=\linewidth]{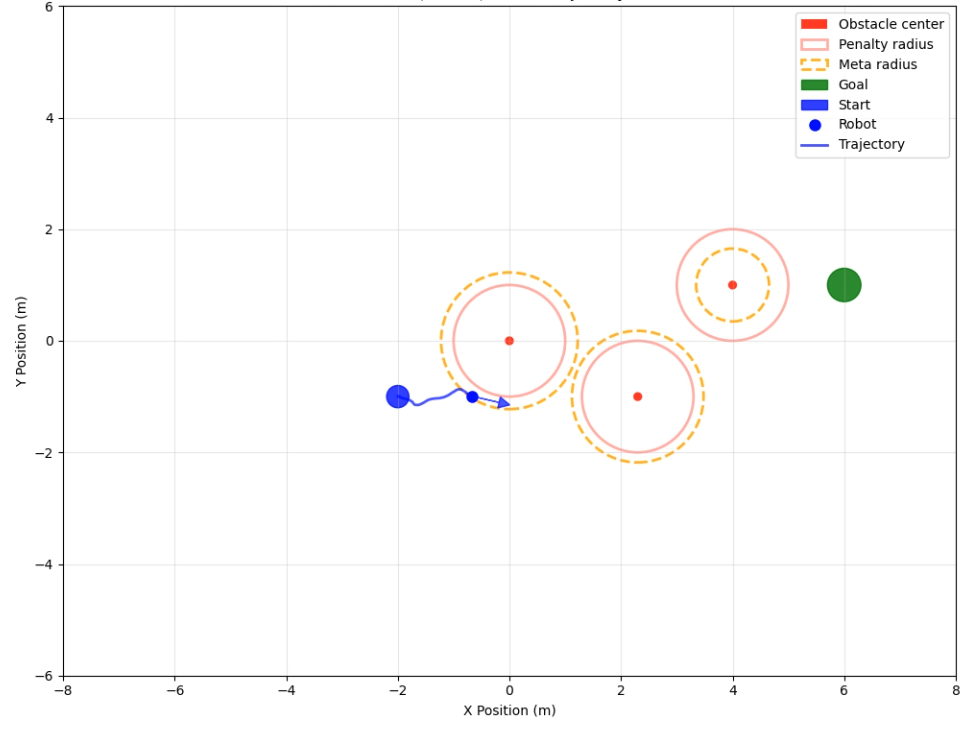}
        \caption{}
    \end{subfigure}
    \hfill
    \begin{subfigure}{0.23\textwidth}
        \centering
        \includegraphics[width=\linewidth]{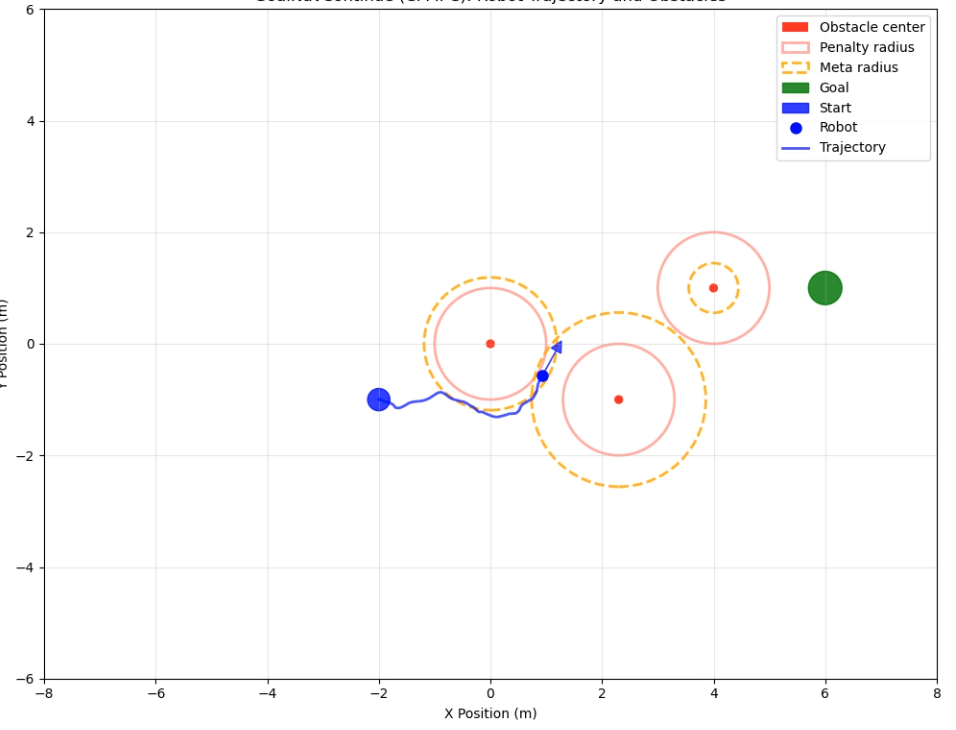}
        \caption{}
    \end{subfigure}

    \vspace{0.2cm}

    \begin{subfigure}{0.23\textwidth}
        \centering
        \includegraphics[width=\linewidth]{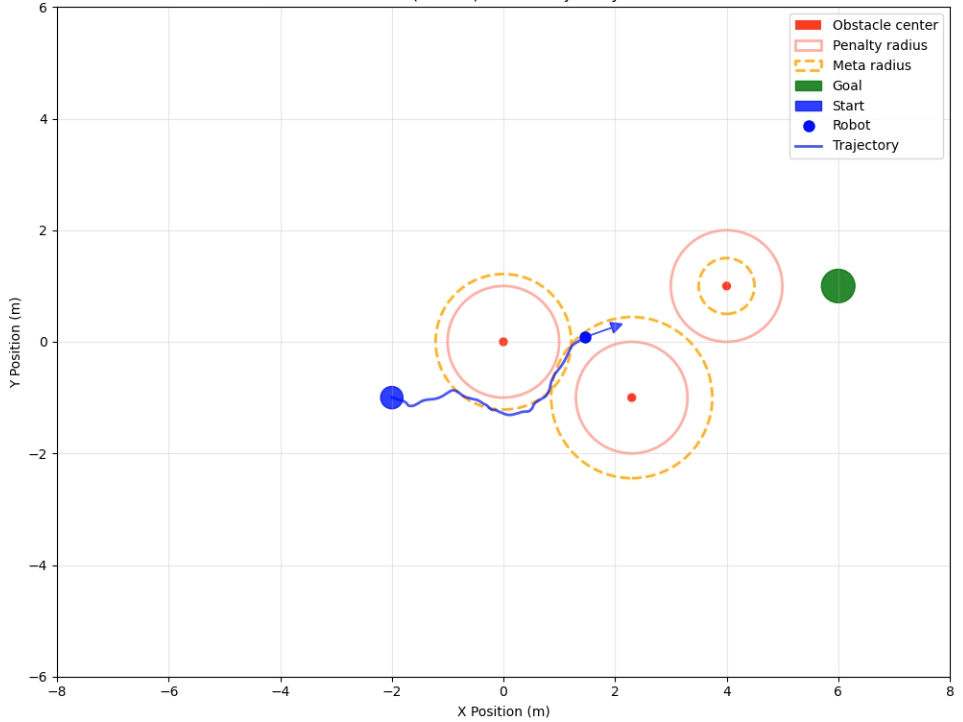}
        \caption{}
    \end{subfigure}
    \hfill
    \begin{subfigure}{0.23\textwidth}
        \centering
        \includegraphics[width=\linewidth]{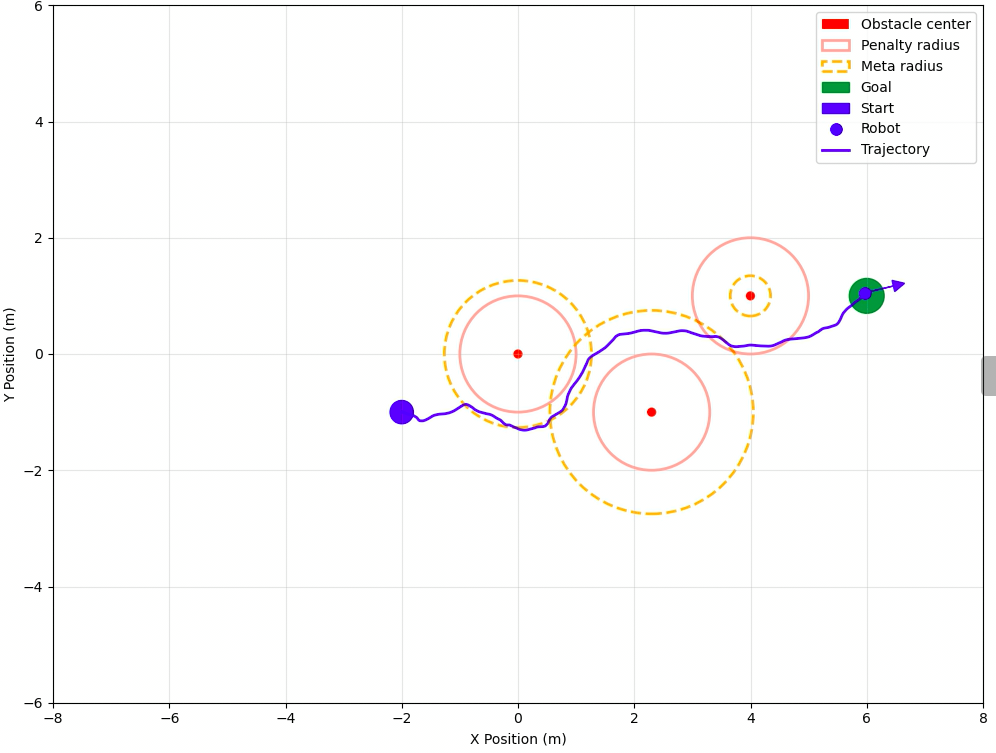}
        \caption{}
    \end{subfigure}

    \caption{Illustration of how RAVEN adapts the planned path. Red dots and solid pink circles denote obstacle centers and physical obstacle radii, respectively. Orange dashed circles denote the adaptive inflation radii generated by the RL meta-policy. The start and goal positions are shown as blue and green dots, respectively, and the robot heading is indicated by a blue arrow.}
    \label{fig:example}
\end{figure}

To illustrate the behavior of RAVEN, we consider a representative navigation scenario, with a delay of 0.06 second. 
The robot starts at the pose $(x,y,\theta)=(3.0,-1.0,-1.57)$ and must reach the goal pose $(6.0,1.8,0.0)$. Three circular obstacles, each with a radius of 1 meter, are centered at $(x,y)=(5.4,-1.4)$, $(4.4,1.0)$, and $(3.0,1.5)$. 
The trajectory is shown in \cref{fig:example}. As the robot approaches the first obstacle, the meta-policy increases the corresponding inflation radius, causing the planner to generate a wider turn and maintain a larger safety margin. When the robot moves through the narrow passage between the first two obstacles, the inflation radii of both obstacles increase, guiding the path along the tangential direction between the inflated circles and preventing collision with the second obstacle. After the robot exits the narrow passage, the inflation radius of the second obstacle decreases, allowing the robot to turn more directly toward the goal pose.

To evaluate the core hypothesis that reinforcement-adapted graph construction improves robustness to real-world uncertainties, we evaluated RAVEN against the baseline methods under varying levels of simulated control delay. Table \ref{tab:delay_metrics} summarizes the performance metrics for a zero-delay ideal scenario and a realistic 0.06~s actuation delay scenario.

In the ideal, zero-delay zero-noise environment (Table \ref{tab:delay_metrics}a), all methods perform competently, though RAVEN already demonstrates superior geometric efficiency, achieving the shortest Path Length and fastest Time to Completion.

However, introducing a 0.06~s delay (Table \ref{tab:delay_metrics}b) drastically alters the performance landscape. The classical DAVG-cfMPC baseline degrades severely; the inability of its static obstacle inflation to account for delayed tracking results in dangerous trajectory overshoot. This causes the Max Obstacle Penetration Depth to spike to 0.128~m, and the reactive avoidance maneuvers balloon the Path Length to 11.25~m.


Conversely, the purely end-to-end RL baseline proves highly resilient to collisions under delay, achieving a remarkable 0.0~m penetration depth. However, this safety comes at the cost of efficiency. Lacking the structural prior of a shortest-path graph, the RL agent adopts an overly conservative, reactive policy that yields a longer average Time to Completion (12.21s) compared to RAVEN.


RAVEN successfully bridges this gap, delivering the best of both paradigms. By observing the delayed system state, the meta-policy dynamically increases the obstacle inflation parameters within the visibility graph. This geometric adaptation naturally guides the underlying MPC along a wider, more conservative trajectory around corners, explicitly mitigating the overshoot caused by the 0.06~s delay without sacrificing global optimality. As shown in the results, RAVEN maintains a safely bounded penetration depth (0.03~m) while achieving the most efficient average path length (9.33~m) and fastest completion time (11.58~s) under delay. This demonstrates that RAVEN outperforms both the brittle static MPC and the overly cautious end-to-end RL agent, preserving smooth, continuous, and efficient trajectories.

\begin{table}[t]
\caption{Simulation Performance Metrics Under Different Delays}
\label{tab:delay_metrics}
\centering
\setlength{\tabcolsep}{6pt}
\renewcommand{\arraystretch}{1.2}

(a) No Delay

\begin{tabular}{lccc}
\hline
Metric & MPC & RL & RAVEN \\
\hline
Average Path Length (m) & 9.838 & 9.645 & \textbf{9.23} \\
Average Time to Completion (s) & 11.43 & 11.55 & \textbf{11.28} \\
Ave. Max Obstacle Penetration Depth (m) & 0.049 &\textbf{0.016} & 0.019 \\
\hline
\end{tabular}

\vspace{0.3cm}

(b) 0.06~s Delay

\begin{tabular}{lccc}
\hline
Metric & MPC & RL & RAVEN \\
\hline
Average Path Length (m) & 11.25 & 9.801 & \textbf{9.33} \\
Average Time to Completion (s) & 12.32 & 12.21 & \textbf{11.58} \\
Ave. Max Obstacle Penetration Depth (m) & 0.128 & \textbf{0.0} & 0.03 \\
\hline
\end{tabular}

\end{table}

\subsection{Hardware Experiments: Sim-to-Real Transfer}
\begin{figure}[h]
    \centering

    \begin{subfigure}{0.98\linewidth}
        \centering
        \includegraphics[width=\linewidth]{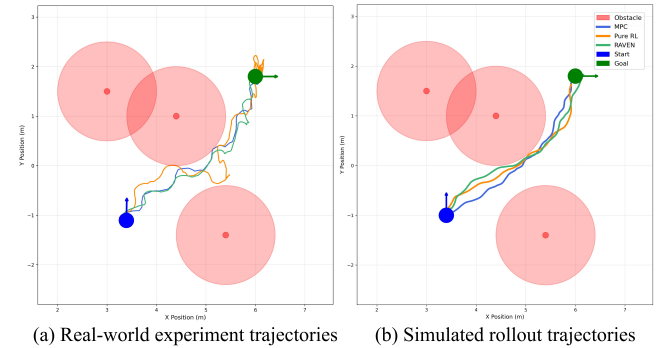}
        \label{fig:sim_traj}
    \end{subfigure}


    \caption{Comparison of real-world and simulated trajectories under the same start pose, goal pose, and obstacle configuration. 
(a) Real-world trajectories recorded by the motion-capture system. 
(b) Simulated rollout trajectories. 
Blue and green dots/arrows denote the start and goal poses, respectively. 
Blue, orange, and green curves denote the DAVG-cfMPC baseline, pure RL baseline, and RAVEN, respectively.}
    \label{fig:trajectory_comparison}
\end{figure}

To validate the practical applicability of the proposed framework, RAVEN was deployed on the T1 humanoid robot from Booster Robotics. For low-level locomotion, we used the open-source Booster Gym locomotion policy~\cite{wang2025booster}. To improve consistency between simulation and hardware, the PyTorch-based locomotion policy was converted into a JIT-compilable format and embedded within our RL-MPC training environment. On the physical robot, the policy was executed locally on the onboard Express-i7 computer, where the low-level joint control loop ran at 500~Hz and policy inference was performed at 50~Hz.

Hardware experiments were conducted on a half-sized RoboCup soccer field. Motion-capture cameras were used to obtain the robot pose and provide localization data for the navigation policy. Due to physical space constraints, mocap coverage was restricted to $[2.0, 6.0]$~m along the $x$-axis and $[-2.5, 2.5]$~m along the $y$-axis; consequently, all start poses, goal poses, and physical obstacles were placed within this mocap area. In both the real-world and simulated experiments, the robot started from $(x,y,\theta)=(3.4,-1.0,1.57)$ and was commanded to reach the goal pose $(6.0,1.8,0.0)$. Three circular obstacles with radius $1.0$~m were placed at $(x,y)=(5.4,-1.4)$, $(4.4,1.0)$, and $(3.0,1.5)$.

The RAVEN planner was executed on an external NVIDIA RTX 4050 GPU and communicated to the T1 robot's onboard computer through ROS2. The DAVG-cfMPC layer alone ran at approximately 120~Hz~\cite{davg-cfmpc}. When combined with the RL meta-policy, the full RAVEN framework ran at 100~Hz, which is sufficient for real-time navigation.

\cref{fig:trajectory_comparison} compares the real-world and simulated trajectories for the same navigation task. 
The pure RL baseline shows a larger discrepancy between simulation and hardware: in the real-world experiment, its trajectory exhibits noticeable side-to-side oscillations that are not present in the simulated rollout.
This suggests that the pure RL policy is more sensitive to the sim-to-real gap. In contrast, the trajectories generated by RAVEN and the DAVG-cfMPC baseline remain relatively consistent between simulation and hardware. These results suggest that the proposed architecture is less sensitive to sim-to-real discrepancies than the pure RL baseline, since the MPC layer provides a model-based control backbone.




\section{Conclusion}

This paper presented \textbf{RAVEN}, a hierarchical RL--MPC framework for robust humanoid navigation. The proposed approach integrates reinforcement learning with classical planning and control by allowing a meta policy to adapt the construction parameters of a visibility-graph planner, while a collision-free MPC controller tracks the resulting trajectory under explicit dynamic and safety constraints. This design enables the system to learn planning adaptations that account for robot dynamics, control delays, and perception noise while preserving the interpretability and safety guarantees of optimization-based control.

Through simulation and hardware experiments, we compared the proposed framework with a manually tuned DAVG--cfMPC baseline and a pure end-to-end RL navigation policy. The results show that RAVEN achieves more reliable navigation in challenging environments, particularly in scenarios involving narrow passages and delayed control. By modifying the geometric structure of the planning graph rather than directly outputting control commands, the RL meta policy can effectively improve robustness while maintaining efficient path planning.

Future work will explore extending the framework to dynamic environments with moving obstacles and integrating more advanced locomotion models to further improve performance on real humanoid platforms.





\end{document}